\documentclass[conference]{IEEEtran}

\usepackage{graphicx}
\usepackage{caption}
\usepackage{subcaption}
\usepackage{rotating}
\usepackage{latexsym}
\usepackage{bbm}
\usepackage{pifont}
\usepackage[nointegrals]{wasysym}
\usepackage{stmaryrd}

\newcommand{\sxrightarrow}[2][]{%
  \mathrel{\text{\footnotesize $\xrightarrow[#1]{#2}$}}%
}

\usepackage{epsfig,amsmath,amsfonts,epsfig,multirow,makecell,caption,soul,csquotes,color,wrapfig,subcaption,mathtools,bm,spverbatim,booktabs,tcolorbox,diagbox}
\usepackage[e]{esvect}

\def\tablename{Table}
\def\figurename{Figure}

\usepackage{caption}
\makeatletter
\newif\if@restonecol
\makeatother

\usepackage[boxed, ruled, vlined, linesnumbered]{algorithm2e}
\SetKwRepeat{Do}{do}{while}

\newenvironment{changemargin}[2]{\begin{list}{}{
	\setlength{\topsep}{0pt}\setlength{\leftmargin}{3pt}
	\setlength{\rightmargin}{0pt}
	\setlength{\listparindent}{\parindent}
	\setlength{\itemindent}{\parindent}
	\setlength{\parsep}{0pt plus 1pt}
	\addtolength{\leftmargin}{#1}\addtolength{\rightmargin}{#2}
	}\item}
	{\end{list}}

\newenvironment{mitemize}{
	\begin{changemargin}{-3pt}{-0cm}
	\vspace{-10pt}
	\hspace{-5pt}
	\begin{itemize}
	\setlength{\itemsep}{3pt}}
	{\end{itemize}
	\vspace{2pt}
	\end{changemargin}}

\newcommand{\ssup}[2]{{#1}^{\scaleobj{0.8}{#2}}}
\newcommand{\ssub}[2]{{#1}_{\scaleobj{0.8}{#2}}}
\newcommand{\sboth}[3]{{#1}_{\scaleobj{0.8}{#2}}^{\scaleobj{0.8}{#3}}}

\usepackage[first=0,last=9]{lcg}
\usepackage{colortbl}
\definecolor{Gray}{gray}{0.8}
\colorlet{Red}{red!10!white}
\colorlet{Blue}{blue!10!white}

\usepackage{hyperref}

\newcommand{\msec}[1]{\S\ref{#1}}
\newcommand{\mref}[1]{\,\ref{#1}}
\newcommand{\meq}[1]{Eq.\,(\ref{#1})}
\newcommand{\mcite}[1]{\,\cite{#1}}

\newcommand{\meg}{\textit{e.g.}\xspace}
\newcommand{\mie}{\textit{i.e.}\xspace}

\newtcolorbox{mtbox}[1]{left=0.25mm, right=0.25mm, top=0.25mm, bottom=0.25mm, sharp corners, colframe=red!50!black, boxrule=0.5pt, title={#1}, fonttitle=\bfseries, coltitle=red!50!black, attach title to upper={\ --\ }}

\usepackage{scalerel}[2016/12/29]

\makeatletter
\providecommand{\leadsfrom}{%
  \mathrel{\mathpalette\reflect@squig\relax}%
}
\newcommand{\reflect@squig}[2]{%
  \reflectbox{$\m@th#1\leadsto$}%
}
\makeatother

\def\eqref#1{equation~\ref{#1}}

\def\1{\bm{1}}

\DeclareMathAlphabet{\mathsfit}{\encodingdefault}{\sfdefault}{m}{sl}
\SetMathAlphabet{\mathsfit}{bold}{\encodingdefault}{\sfdefault}{bx}{n}

\def\gA{{\mathcal{A}}}

\def\gG{{\mathcal{G}}}

\def\gK{{\mathcal{K}}}
\def\gL{{\mathcal{L}}}

\def\gR{{\mathcal{R}}}
\def\gS{{\mathcal{S}}}

\def\gV{{\mathcal{V}}}

\begin{document}

\newcommand{\kg}{{KG}\xspace}
\newcommand{\kgs}{{KGs}\xspace}

\newcommand{\qa}{{\sc QA}\xspace}
\newcommand{\cve}{{CVE}\xspace}
\newcommand{\rec}{{\sc Rec}\xspace}
\newcommand{\krl}{{KRL}\xspace}
\newcommand{\mrr}{{MRR}\xspace}
\newcommand{\hit}{{HIT@$K$}\xspace}
\newcommand{\hito}{{HIT@$1$}\xspace}

\newcommand{\system}{{\sc RoMA}\xspace}

\newcommand{\NA}{----}

\newcommand{\vect}[1]{\boldsymbol{#1}} 

\newcommand{\ting}[1]{\textcolor{purple}{[#1]}\xspace}

\newcommand{\zhaohan}[1]{\textcolor{blue}{{#1}}\xspace}

\title{Reasoning over Multi-view Knowledge Graphs}
\author{
{\rm Zhaohan Xi}$^\dagger$ \,\, {\rm Ren Pang}$^\dagger$ \,\, {\rm Changjiang Li}$^\dagger$ \,\, {\rm Tianyu Du}$^\dagger$ \,\, {\rm Shouling Ji}$^\ddagger$ \,\, {\rm Fenglong Ma}$^\dagger$ \,\, {\rm Ting Wang}$^\dagger$\\
$^\dagger$Pennsylvania State University \quad
$^\ddagger$Zhejiang University and Ant Financial\\
} 

\maketitle
\thispagestyle{empty}

\begin{abstract}

Recently, knowledge representation learning (KRL) is emerging as the state-of-the-art approach to process queries over knowledge graphs (KGs), wherein KG entities and the query are embedded into a latent space such that entities that answer the query are embedded close to the query. Yet, despite the intensive research on KRL, most existing studies either focus on homogenous KGs or assume KG completion tasks (\mie, inference of missing facts), while answering complex logical queries over  KGs with multiple aspects (multi-view KGs) remains an open challenge.

To bridge this gap, in this paper, we present \system\footnote{\system: \underline{R}easoning \underline{O}ver \underline{M}ulti-view Knowledge Gr\underline{A}phs.}, a novel KRL framework for answering logical queries over multi-view KGs. Compared with the prior work, \system departs in major aspects. (i) It models a multi-view KG as a set of overlaying sub-KGs, each corresponding to one view, which subsumes many types of KGs studied in the literature (\meg, temporal KGs). (ii) It supports complex logical queries with varying relation and view constraints (\meg, with complex topology and/or from multiple views); (iii) It scales up to KGs of large sizes (\meg, millions of facts) and fine-granular views (\meg, dozens of views); (iv) It generalizes to query structures and KG views that are unobserved during training.  Extensive empirical evaluation on real-world KGs shows that \system significantly outperforms alternative methods.

{\em Keywords --} Knowledge Graph; Representation Learning

\end{abstract}

\section{Introduction}
\label{sec:intro}

\begin{figure}[!t]
    \centering
    \epsfig{file = 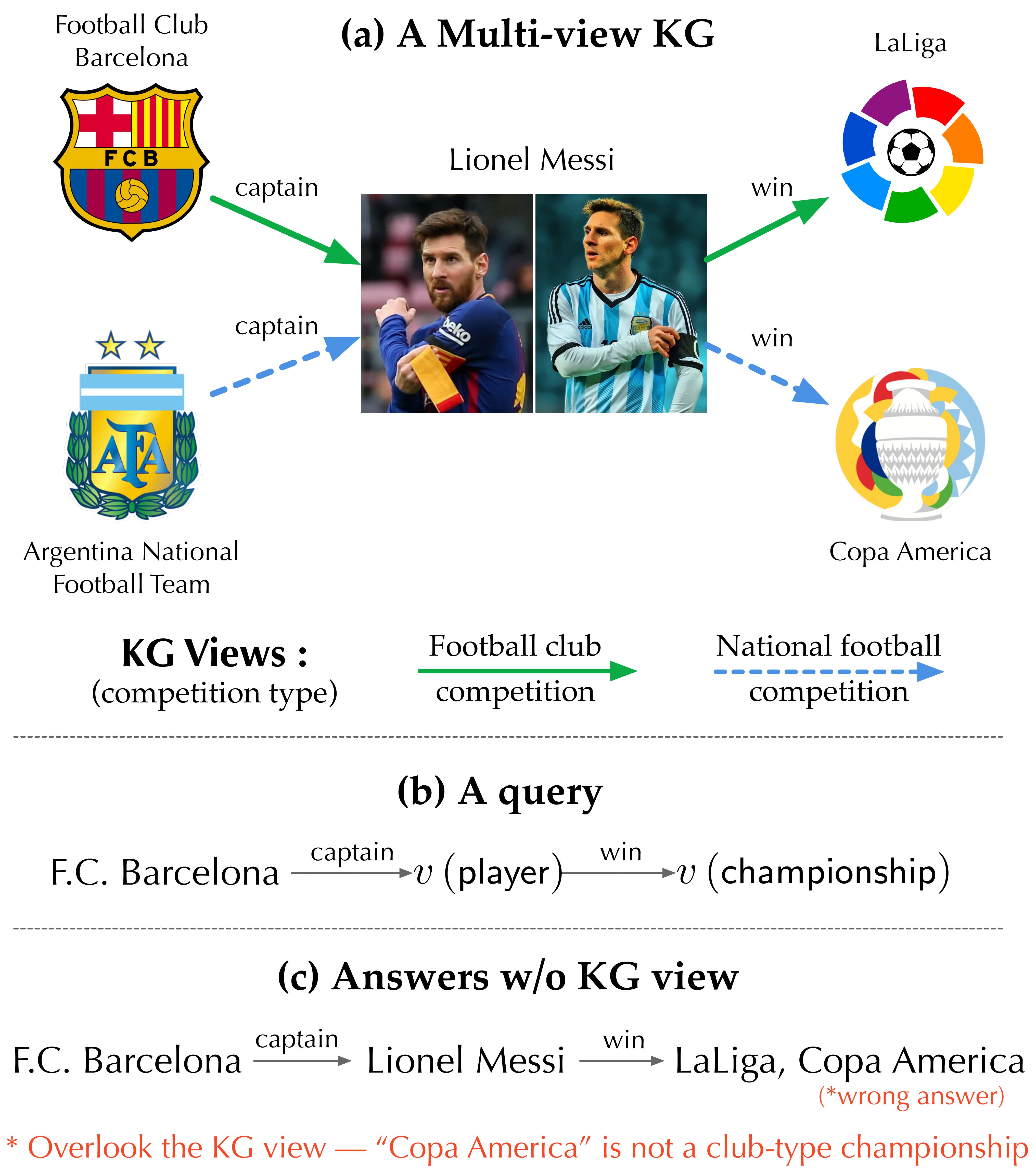, width = 88mm}
    \caption{A motivating example of query answering over multi-view KGs (each sport season as one view), where the view constraint is critical to ensure the answer correctness.}
    \label{fig:example}
\end{figure}

A knowledge graph (\kg) is a structured representation of human knowledge, with entities, relations, and descriptions respectively capturing real-world objects (or abstract concepts), their relationships, and their semantic properties. For example, Figure\mref{fig:example}(a) describes the relationships among a few football players, clubs, and championships.
Answering complex logical queries over \kgs represents an important artificial intelligence task, entailing a wide range of applications. 


Recently, knowledge representation learning (\krl) is emerging as the state-of-the-art approach to reason over \kgs, wherein \kg entities and a query are projected into a latent space such that the entities that answer the query are embedded close to each other. 
Essentially, \krl reduces answering an arbitrary query to simply identifying entities with embeddings most similar to the query, thereby implicitly imputing missing relations\mcite{missing-relation} and scaling up to large-scale {\kgs}\mcite{yago}.

Yet, despite the intensive research on {\krl}\mcite{transe, transr, rotate, iclr-kge-tricks,logic-query-embedding, query2box, beta-embedding}, most existing work focuses on capturing the relational structures of \kgs while ignoring the fact that each \kg often comprises multiple aspects. For instance, a multilingual \kg is a collection of sub-KGs, each corresponding to one language\mcite{cross-lingual-kg, multilingual}. We refer to such sub-KGs as ``views'', analogous to the view concept in the computer vision domain\mcite{multi-view-survey}. Intuitively, each view of a KG reflects one specific aspect of the knowledge. We refer to KGs with multiple views as multi-view KGs. For example, Figure\mref{fig:example} shows a multi-view KG with two views, each corresponding to one particular competition.

It is often essential to account for the view information in answering queries over multi-view KGs. Consider the logical query as shown in Figure\mref{fig:example}(b): 
\begin{quote}
\centering
``{\em Which championship} did the {\em football player} win when playing as the captain of \textsf{\small F.C. Barcelona}?'' 
\end{quote}
Without view constraints, as shown in Figure\mref{fig:example}(c), KRL may reach the answer set of $\{\textsf{\small LaLiga, Copa America}\}$ via traversing the KG, which is erroneous as the relations of $\textsf{\small F.C. Barcelona} \sxrightarrow[]{\text{captain}} \textsf{\small Lionel Messi} $ and $\textsf{\small Lionel Messi} \sxrightarrow[]{\text{win}} \textsf{\small Copa America} $ do not exist in a same competition. 

Note that recent work also studies temporal KGs: a sequence of KG snapshots represent its temporal evolution\mcite{chronor, temporal-krl, temporal-box-embed, kdd-temporal-kgc, temporal-kg-encode}. However, temporal KGs are a special class of multi-view KGs in which the view is defined as the {\em timestamp} of each KG snapshot. Unlike temporal KGs, general multi-view KGs do not assume a sequential order among different views. Thus, answering complex queries over general multi-view KGs requires to traverse all the views simultaneously. Moreover, the query structures may scatter across different views. Hence, designing novel \krl to support reasoning over general multi-view KGs remains an open challenge.

Thus, in this work, we formalize the task of {\em answering queries over multi-view KGs}, which 
entails the following challenges:

\vspace{2pt}
\noindent
Q$_1$ -- How to represent general multi-view \kgs in \krl?

\vspace{2pt}
\noindent
Q$_2$ -- How to enforce the view constraints in query answering?

\vspace{2pt}
\noindent
Q$_3$ -- How to generalize to the views unobserved in training?

\vspace{4pt}
\textbf{This work --} We present \system, a novel KRL framework for answering logical queries over multi-view \kgs, which tackles the aforementioned challenges as follows:

\vspace{2pt}
RA$_1$ -- \system encodes different views of \kgs into a unified latent space, wherein the embeddings of KG entities capture two aspects of information: the entities' relation structures and their existence in different views. Note that \system does not assume specific embedding definitions (\meg, vectors, boxes, or distributions) and supports varying embedding geometries (details in \msec{ssec:geo}).

\vspace{2pt}
RA$_2$ -- \system supports complex logical queries with varying relation (\meg, with complex topology) and view  (\meg, from multiple views) constraints. Specifically, it employs two decoders respectively responsible for decoding relation and view information; at each iteration, it runs the two decoders in parallel to enforce the relation and view constraints and merges the intermediate results; it is thus able to derive high-quality answers that meet both relation and view constraints. 


\vspace{2pt}
RA$_3$ -- \system readily generalizes to query structures and KG views that are not observed during training. Specifically, given its modular design of query answering (\mie, iterative application of basic operators), it is able to handle queries with unseen relation structures; meanwhile, due to its positional encodings of views (details in \msec{ssec:coding}), it is able to interpolate or extrapolate views that are unobserved during training.


\vspace{4pt}
\textbf{Contributions --} To our best knowledge, this work represents the first study on answering logical queries over multi-view KGs. Our contributions can be summarized as follows.

\vspace{2pt}
-- We formalize the concept of multi-view KGs which subsumes many existing types of KGs (\meg, multi-lingual, temporal KGs), and define the task of answering logic queries over multi-view KGs.

\vspace{2pt}
-- We present \system, a novel KRL framework for multi-view KGs, which effectively encodes both relation and view information and supports queries with varying relation structures and view dependencies.

\vspace{2pt}
-- We conduct extensive evaluation of \system and demonstrate its reasoning capability by comparing it with existing temporal KRL methods, exploring different embedding geometries, and characterizing its training dynamics.

\vspace{4pt}
\textbf{Roadmap --} The remainder of the paper will proceed as follows: \msec{sec:background} defines the task of answering logical queries over multi-view KGs; \msec{sec:tech} details the design of \system;  \msec{sec:expt} conducts an empirical evaluation of \system on real-world KGs; \msec{sec:literature} surveys related literature; the paper is concluded in \msec{sec:conclusion}.
\section{Preliminaries}
\label{sec:background}

\begin{table}[!h]{\small 
\centering
\renewcommand{\arraystretch}{1.2}
\begin{tabular}{c | l }
Notation & Definition \\
\hline
\hline
$\gG = (\gV, \gR)$ & knowledge graph = (entities, relations)\\
$\vartheta \in \Theta$  & view $\in$ views \\
$q = \langle v_?, \gK_q, \gV_q, $  & query = $\langle$entity of interest, anchors, variables,\\ $\gR_q, \Theta_q \rangle$ &  relation constraints, view constraints$\rangle$ \\
$\llbracket q \rrbracket$, $\gA_q$ & query answers, candidate answers\\
$e_v$, $\theta_v$ & entity $v$'s semantic encoding, view encoding
\end{tabular}
\caption{Notations and definitions. \label{tab:notation}}}
\end{table}

In this section, we introduce fundamental concepts and assumptions, formally define the problem of reasoning over multi-view KGs, and then discuss alternatives to motivate the design of \system. Table\mref{tab:notation} summarizes the key notations and definitions.


\subsection{Multi-view Knowledge Graphs}

We define a knowledge graph (KG) as $\gG = (\gV, \gR)$, where $v \in \gV$ denotes an entity and $r \in \gR$ is a binary function $r: \gV \times \gV \rightarrow \{\text{True}, \text{False}\}$ indicating whether there exists relation $r$ between two entities. That is, there is a directed edge (fact) $v \sxrightarrow{r} v'$ in $\gG$ between 
$v, v' \in \gV$ iff $r(v, v') = \text{True}$. 



Multi-view \kgs extend the above definition to provide finer-grained representations. Specifically, a multi-view KG $\gG$ is defined as a set of overlaying sub-KGs $\gG = \{ \gG_\vartheta \}_{\vartheta \in \Theta}$, where $\Theta$ denotes a set of ``views'', each corresponding to one specific aspect of $\gG$, while each sub-KG $\gG_\vartheta = (\gV, \gR_\vartheta)$ shares the same set of entities with $\gG$ but has its unique relations. For instance, the multi-view KG in Figure\mref{fig:example} comprises three sub-KGs, each corresponding to one sport season. Note that the definition of multi-view KGs subsumes many types of KGs. For instance, temporal KGs\mcite{temporal-kg-encode, know-evolve} are special cases of multi-view KGs wherein each view corresponds to one specific time, while in multi-linguistic KGs, each language can be considered as a specific view\mcite{cross-lingual-kg}.


\subsection{Logical Query Answering}
\label{ssec:qa}

A variety of reasoning tasks can be performed over KGs. For instance, given that KGs are often incomplete, KG completion aims to infer missing relations\mcite{chains-of-reasoning,path-query}. Here, we focus on more complex queries that ask for specific entities satisfying given logical constraints. As an example, 
\begin{quote}
``{\em Where} did \textsf{\small Messi} play as a football team captain?'' 
\end{quote}
Such queries are often defined using first-order logic with conjunctive ($\wedge$) and existential ($\exists$) operations. Formally, a conjunctive query $q$ comprises $v_?$ as the entity of interest, $\gK_q$ as a set of known entities ({\em anchors}), $\gV_q$ as a set of existentially quantified bounded entities ({\em variables}), and $\gR_q$ as a set of relations. The answer to $q$ is denoted by $\llbracket q \rrbracket$. For instance, the above query can be formulated as: 
\begin{equation*}
    q = v_? . \exists v: (\textsf{\small Messi} \sxrightarrow[]{\text{captain of}} v)\wedge (v \sxrightarrow[]{\text{locate at}} v_?)
\end{equation*}
where $\gK_q$ = $\{ \textsf{\small  Messi}\}$, $\gV_q$ = $\{v\,(\text{football team}) \}$, $\gR_q$ = $\{
\textsf{\small Messi} \sxrightarrow[]{\text{play for}} v, 
v \sxrightarrow[]{\text{locate at}} v_?
\}$, and $\llbracket q \rrbracket = \{\textsf{\small Barcelona}, \textsf{\small Argentina}\}$.

\vspace{2pt}
Essentially, query processing over \kgs is a graph matching problem that matches the entities and relations of query graphs with that of \kgs. However, its computational complexity tends to grow exponentially with the query size; real-world \kgs often contain missing relations\mcite{missing-relation}, further impeding exact graph matching. Recently, knowledge representation learning (\krl) has emerged as a promising approach\mcite{query2box, beta-embedding, ren2021lego}, wherein \kg entities and a query are projected into a latent space such that the entities that answer the query are embedded close to each other. \krl reduces answering an arbitrary query to simply identifying entities with embeddings most similar to the query, thereby implicitly imputing missing relations and scaling up to large-scale \kgs.

Specifically, to process a conjunctive query $q$ with respect to KG $\gG$, one derives $q$'s computation graph from $\gG$ with two operators: projection - given entity set $\gS$, it computes the entities with relation $r$ to $\gS$; intersection - given entity sets $\{\gS_i\}_i$, it computes their intersection $\cap_i \gS_i$. Intuitively, $q$'s computation graph specifies the procedure of answering $q$ as traversing $\gG$: starting from  anchors $\gK_q$, it iteratively applies the two operators to compute variables $\gV_q$, until reaching the answer set $\llbracket q \rrbracket$. For instance, in the above example, starting with $\gK_q=\{\textsf{\small Messi}\}$, it follows relations $\gR_q$ to arrive at variables $\gV_q=\{\textsf{\small F.C. Barcelona}, \textsf{\small Argentina National Team}\}$ and eventually reaches $\llbracket q \rrbracket=\{\textsf{\small Barcelona}, \textsf{\small Argentina}\}$.

\begin{figure*}[!t]
    \centering
    \epsfig{file = 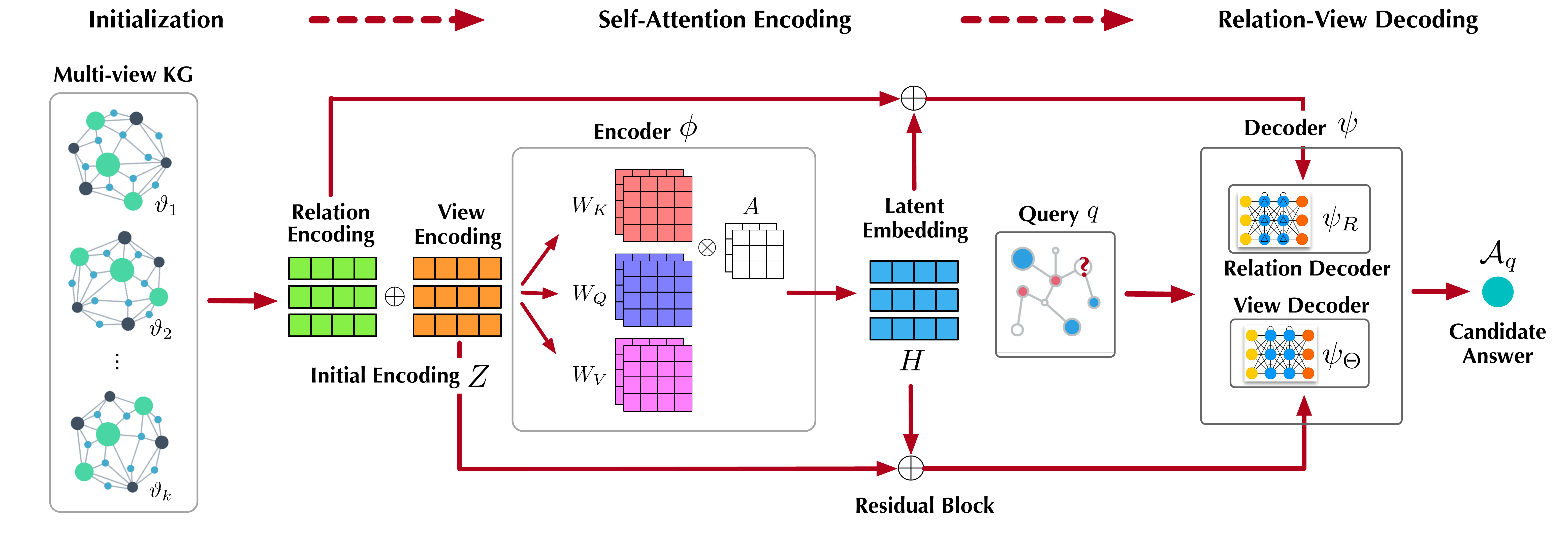, width = 185mm}
    \caption{Overview of \system.}
    \label{fig:design}
\end{figure*}

\subsection{Query Answering over Multi-view KGs}
\label{ssec:multi-version-qa}

On multi-view KGs, we extend a conjunctive query $q$ with view constraints $\Theta_q$ as $q = \langle v_?, \gK_q, \gV_q , \gR_q, \Theta_q \rangle$, in which $\Theta_q$ allows to specify various constraints for relation $r \in \gR_q$:
\begin{mitemize}
\item Exact match: $\vartheta(r) = \vartheta^*$ specifies that $r$ must appear in view $\vartheta^*$ (\mie, sub-KG $\gG_{\vartheta^*}$);
\item Wildcard match: $\vartheta(r) = *$ specifies that there is no view constraint imposed on $r$; 
\item Equal match: $\vartheta(r) = \vartheta(r')$ specifies that $r, r' \in \gR_q$ must appear in the same view.
\end{mitemize}

For example, consider a query
\begin{equation*}
    q = v_? . \exists v: (\textsf{\small F.C. Barcelona} \sxrightarrow[]{\text{captain}} v) \wedge (v \sxrightarrow[]{\text{win}} v_?)
\end{equation*}
which asks for the championship $v_?$ a football player $v$ has won when playing as the captain of the football club $\textsf{\small F.C. Barcelona}$. Without the view constraints, $\llbracket q \rrbracket$ contains $\textsf{\small Copa America}$, because of the logical path $\textsf{\small F.C. Barcelona} \sxrightarrow[]{\text{captain}} \textsf{\small Messi} \sxrightarrow[]{\text{win}} \textsf{\small Copa America}$ in $\gG$; yet, with the equal match constraint 
$\vartheta(\textsf{\small F.C. Barcelona} \sxrightarrow[]{\text{captain}} v) = \vartheta(v \sxrightarrow[]{\text{win}} v_?)$, $\textsf{\small Copa America}$ is not a correct answer, given that $\textsf{\small Copa America}$ a national competition while  $\textsf{\small F.C. Barcelona}$ is only able to participate in club-level competitions.

Intuitively, as answering queries over multi-view KGs needs to satisfy both relation constraints and view constraints, such finer-granular constraints further aggravate the challenges of computational complexity and KG incompleteness, thereby demanding for new query processing techniques.

\subsection{Alternative Solutions}
\label{ssec:alternative}

Here, we consider alternative solutions for reasoning over multi-view KGs, which motivate the design of \system.

One straightforward approach is to consider a multi-view KG $\gG$ as static (\mie, ignoring its view information), apply existing KRL methods to generate entity embeddings, and filter out variables that fail to meet the view constraints during query processing. Because of its view-agnostic nature, this approach readily supports queries with varying view constraints but meanwhile suffers a high false-positive rate and is sensitive to the issue of KG incompleteness. 

In contrast, another approach is to apply KRL to each sub-KG $\gG_\vartheta$ to generate view-specific entity embeddings. This approach is highly effective to process queries with specific match constraints. However, for non-specific view constraints (\meg, wildcard or equal match), due to its lack of cross-view interoperability, this approach often needs to enumerate all possible views, impacting its efficiency. For example, to process query $q = v_?.\exists v: (\textsf{\small F.C. Barcelona} \sxrightarrow[]{\text{captain}} v) \wedge (v \sxrightarrow[]{\text{win}} v_?)$ with equal match, one needs to enumerate all the views to check whether $\textsf{\small F.C. Barcelona} \sxrightarrow[]{\text{captain}} v$ and $v \sxrightarrow[]{\text{win}} v_?$ both hold. This complexity grows sharply with the query size.

Next, we present \system, a view-aware KRL method for reasoning over multi-view KGs, which enjoys the strengths of the above solutions without sharing their limitations.

\section{Design of ROMA}
\label{sec:tech}

\subsection{Overview}

At a high level, \system follows an encoder-decoder framework\mcite{dynamic-gnn-survey} as illustrated in Figure\mref{fig:design}: given the initial encodings of KG entities, the encoder $\phi$ projects \kg entities into a latent space as embeddings, while the decoder $\psi$ takes as input the embeddings of KG entities and queries to perform logical reasoning. 

More specifically, the encoder applies the self-attention mechanism and encodes the relational structures of KG entities as well as their view dependencies; the decoder comprises two sub-decoders, structure decoder and view decoder, which naturally enforce relation constraints and view constraints during query processing.

Below we elaborate on each key component.





\subsection{Initialization}
\label{ssec:coding}

In the initialization stage, \system generates the initial encodings of KG entities. To overcome the drawbacks of view-agnostic and view-specific encodings in \msec{ssec:alternative}, we propose the following design.



Strictly speaking, each view $\vartheta$ is associated with the relations of KG $\gG$ in sub-KG $\gG_\vartheta = (\gV, \gR_\vartheta)$. Yet, to facilitate encoding KG entities, we consider views as additional attributes of entities. In other words, if relation $v \sxrightarrow[]{r} v'$ appears in view $\vartheta$, then $\vartheta$ is an attribute of both $v$ and $v'$. This relaxation significantly simplifies the encodings, at a slight cost of false-positive rate. 

Specifically, the initial encoding of each entity $v \in \gV$ consists of two parts: (i) the $d$-dimensional semantic encoding $e_v \in \mathbbm{R}^d$ capturing $v$'s relational structure within $\gG$, which is randomly initialized and optimizable during model training; (ii) the $d$-dimensional view encoding $\theta_v \in \mathbbm{R}^d$ capturing $v$'s presence in different views.

To compute $\theta_v$, we first assign a fixed (untrainable) encoding to each view $\vartheta \in \Theta$. A variety of designs are possible. In the case that the views are uncorrelated, simple one-hot encoding is applicable; in the case that the views are ordered, positional encoding preserves the ordering information. In our current implementation, we use positional encoding based on trigonometric functions, which is often used in Transformer models\mcite{attn-is-all-you-need}. Let $\textsf{PosEnc}(\vartheta)$ denote $\vartheta$'s positional encoding.

We then aggregate the set of views $\Theta_v$ associated with $v$. To this end, we apply a permutation-invariant neural network $\textsf{SetEnc}$\mcite{deepset} as the aggregation function for the set of encodings of $\Theta_v$ to generate the view-set encoding $\theta_v$, which is invariant to the order of views in $\Theta_v$. Formally,
\begin{equation}
\label{eq:verenc}
\theta_v = \textsf{SetEnc}(\{\textsf{PosEnc}(\vartheta) \;|\; \vartheta \in \Theta_v \})
\end{equation}


For each entity $v \in \gV$, given its semantic encoding $e_v$ and view-set encoding $\theta_v$, we further generate $z_v = e_v \oplus \theta_v$ via  element-wise addition, as $v$'s initial encoding fed to the encoder $\phi$.

\subsection{Self-Attention Encoding}
\label{ssec:encode}

The encoder $\phi: \mathbbm{R}^{|\gV| \times d} \rightarrow  \mathbbm{R}^{|\gV| \times d}$ maps the initial encodings of KG entities $Z \in \mathbbm{R}^{|\gV| \times d}$ (in a matrix form) to their latent embeddings $H \in \mathbbm{R}^{|\gV| \times d}$, in which each row corresponds to one distinct KG entity. For simplicity, here we assume $Z$ and $H$ share the same dimensionality.

To make $\phi$ capture the topological structures of the KG, we adopt the self-attention mechanism, which is widely used to compute the pair-wise importance between words (without explicitly providing their interconnections)\mcite{attn-is-all-you-need, gat}. Given the initial encodings $Z$ and adjacency matrix $A$ (derived from the KG), the self-attention layer of $\psi$ employs two shared linear projection matrices $\ssub{W}{Q}, \ssub{W}{K}$ to compute a coefficient matrix $C$, representing the pair-wise importance of entities masked by $A$:
\begin{equation}
\label{eq:attn:coefficient}
\begin{gathered}
    C = \textsf{Attn}(Q, K, A) = \textsf{softmax}\left(\frac{Q K^{\top}}{\sqrt{d}}\right) \otimes A\\
    \text{s.t.}\quad Q=Z \ssub{W}{Q}, \,\, K=Z\ssub{W}{K} 
\end{gathered}
\end{equation}
Then, $\phi$ updates $Z$ by another projection matrix $\ssub{W}{V}$:
\begin{equation}
\label{eq:attn:update}
H = \textsf{Update} (C, V) = CV \quad  \text{s.t.} \quad V = Z\ssub{W}{V}
\end{equation}
where $H$ represents the embeddings of KG entities and is later used by the decoder $\psi$ for query processing (details in \msec{ssec:decode}).

Combining \meq{eq:attn:coefficient} and \meq{eq:attn:update}, we define the encoder $\phi$ as:
\begin{equation}
\label{eq:encoder}
H = \phi(Z) \quad \text{s.t.} \quad \phi \supseteq \{\textsf{Attn}, \textsf{Update}\}
\end{equation}


Note that the masking matrix $A$ is agnostic to relation types. We may further account for relation types and construct relation-specific masking matrices to replace $A$ in \meq{eq:attn:coefficient}. The relation-specific attention is similar to multi-head attention but differs in using different adjacency matrices to mask entities.


\subsection{Relation-View Decoding}
\label{ssec:decode}

The decoder $\psi$ comprises two sub-decoders $\ssub{\psi}{R}$, $\ssub{\psi}{\Theta}$, which run in parallel to enforce a given query $q$'s relation and view constraints. To make $\ssub{\psi}{R}$ and $\ssub{\psi}{\Theta}$ respectively focus on matching relation and view constraints, we add residual blocks that combine the initial relation and view encodings with the embeddings of KG entities as inputs to $\ssub{\psi}{R}$ and $\ssub{\psi}{\Theta}$, as illustrated in Figure\mref{fig:design}. In \msec{ssec:ablation}), we empirically evaluate the impact of this design on the reasoning efficacy.


Following the typical design of KRL\mcite{query2box, hamilton2018embedding, beta-embedding}, $\ssub{\psi}{R}$ comprises a set of neural networks that implement the projection (specific for each relation type) and intersection operators as introduced in \msec{ssec:qa}. Similarly, $\ssub{\psi}{\Theta}$ matches $q$'s view constraints with $\gG$ using a set of neural networks that implement the projection and intersection operators. Each operator is specific for one type of view constraints (\meg, equal or wildcard match). Besides, we define a new {\em merger} operator that merges the intermediate results returned by  $\ssub{\psi}{R}$ and $\ssub{\psi}{\Theta}$ during the reasoning process. 

To perform reasoning, starting with $q$'s anchors $\gK_q$, \system extracts their embeddings from $H$, and iteratively applies the projection and intersection operators of $\ssub{\psi}{R}$ and $\ssub{\psi}{\Theta}$ following $q$'s structure; at each step, it applies the merger operator to merge the intermediate results (\meg, boxes in\mcite{query2box}) of $\ssub{\psi}{R}$ and $\ssub{\psi}{\Theta}$, which serves as the starting point for $\ssub{\psi}{R}$ and $\ssub{\psi}{\Theta}$ for the next step; after deriving the embeddings of query answers $\llbracket q \rrbracket$ denoted as $h_{\llbracket q \rrbracket}$, \system matches $h_{\llbracket q \rrbracket}$ with $H$ to identify the candidate entities, denoted by $\gA_q$. This decoding procedure is sketched in Algorithm\mref{alg:decode}.

We measure the quality of the match between $\llbracket q \rrbracket$ and a candidate answer $a \in \gA_q$ from both relation and view perspectives. On the relation end, we apply a score function similar to\mcite{query2box, hamilton2018embedding, beta-embedding} to compute the latent-space similarity between $\llbracket q \rrbracket$ and $a$. Letting $\sboth{h}{a}{(R)}$ and $\sboth{h}{\llbracket q \rrbracket}{(R)}$ be $a$'s and $\llbracket q \rrbracket$'s embeddings in the relation latent space, respectively, we compute the score of $a$ as:
\begin{equation}
\label{eq:struc:score}
\ssub{\textsf{Sim}}{R}(a) = \gamma - \textsf{Dist}
\left(\sboth{h}{\llbracket q \rrbracket}{(R)}, \sboth{h}{a}{(R)}
\right)
\end{equation}
where $\gamma$ is a given scalar margin and $\textsf{Dist}(\cdot, \cdot)$ is the distance metric (\meg, $L_p$ norm). Similarly, on the view end, we define a score function to measure the similarity of $\llbracket q \rrbracket$ and $a$, with a higher score indicating the co-existence of $a$ and $\llbracket q \rrbracket$ in more views. Concretely, we use element-wise correlation as the metric:
\begin{equation}
\label{eq:ver:score}
\ssub{\textsf{Sim}}{\Theta}(a) = \frac{1}{d} \sum\left(\sboth{h}{\llbracket q \rrbracket}{(
\Theta)} \otimes \sboth{h}{a}{(\Theta)}\right)
\end{equation}
where $\sboth{h}{a}{(\Theta)}$ and $\sboth{h}{\llbracket q \rrbracket}{(\Theta)}$ are $a$'s and $\llbracket q \rrbracket$'s embeddings in the view latent space.


We rank the candidate answers based on their overall scores 
$\textsf{Sim}(a) = \ssub{\textsf{Sim}}{R}(a) \cdot \ssub{\textsf{Sim}}{\Theta}(a)$ for $a \in \gA_q$ and select the top candidates as the final answers (see Algorithm\mref{alg:decode}). In \msec{ssec:train}, we also define the training objective based on the overall scores.

\begin{algorithm}[!t]{\small 
\KwIn{
    $H$ -- latent embeddings; 
    $q = \langle v_?, \gK_q, \gV_q , \gR_q, \Theta_q \rangle$ -- query;
    $\gA_q$ -- candidate entities;
    $n$ -- number of entities to retrieve;
    $\ssub{\psi}{R}, \ssub{\psi}{\theta}$ -- relation decoder and view decoder
}
\KwOut{
    $\llbracket q \rrbracket$ -- answers to $q$
}
\tcp{initialization}
$\gL \leftarrow \emptyset$\;
\ForEach{$v \in \gK_q$}{
extract $h_v$ from $H$ and add $h_v$ to $\gL$ \tcp*{initialize queue}
}
\tcp{query traversal}
\While {$\gL \neq \emptyset$} {
   dequeue $h$ from $\gL$ \tcp*{fetch current embedding}
   follow $\gR_q$ and apply $\ssup{h}{(R)} \leftarrow \ssub{\psi}{R}(h)$ \tcp*{relation constraint}
   follow $\Theta_q$ and apply  $\ssup{h}{(\Theta)} \leftarrow \ssub{\psi}{\Theta}(h)$  \tcp*{view constraint}
   merge  $\ssup{h}{(R)}$ and $\ssup{h}{(\Theta)}$ as updated $h$ \tcp*{merger operator}
 \lIf{$\gL \neq \emptyset$}{
add $h$ to $\gL$}
}

    
\tcp{answer matching}
match $h$ with $H$ to identify candidates $\gA_q$\;
\ForEach{$a \in \gA_q$}{
    compute $\textsf{Sim}(a)$    \tcp*{\meq{eq:struc:score} and \meq{eq:ver:score}}
}
\Return top-$n$ entities of $\gA_q$ as $\llbracket q \rrbracket$\;
\caption{Relation-View Decoding \label{alg:decode}}}
\end{algorithm}

\subsection{End-to-End Optimization}
\label{ssec:train}

Within the encoder-decoder framework, we train \system (e in an end-to-end manner by feeding it with the \kg and sampled queries (with answers) and maximizing the likelihood of finding ground-truth answers that meet both relation and view constraints. 


Specifically, to construct the training set, we apply the following sampling procedure: we first randomly sample a view constraint (\mie, exact, wildcard, or equal match); under this constraint, we sample a query $q$ from the KG; we then apply negative sampling to pair $q$ with a ground-truth answer $a \in \llbracket q \rrbracket$ and $k$ negative answers $\{\bar{a}_i | \bar{a}_i 
\not \in \llbracket q \rrbracket\}_{i=1}^k$. We define the loss with respect to $q$ as:
\begin{equation}
\label{eq:loss}
\ell(q) = -\textsf{log}\sigma(\textsf{Sim}(a) ) - \frac{1}{k} \sum_{k}^{i=1} \textsf{log}\sigma(- \textsf{Sim}(\bar{a}_i) )
\end{equation}

At each training step, the average loss over a batch of query-answer samples $\left<q, a, \{\bar{a}\}\right>$ is computed and backward propagated to update the decoder $\psi=\{\ssub{\psi}{R}, \ssub{\psi}{\Theta} \}$, the encoder $\phi$, the initial semantic encodings $\{e_v\}_{v \in \gV}$, and the view-set aggregator $\textsf{SetEnc}$.

\begin{figure*}[!ht]
    \centering
    \epsfig{file = 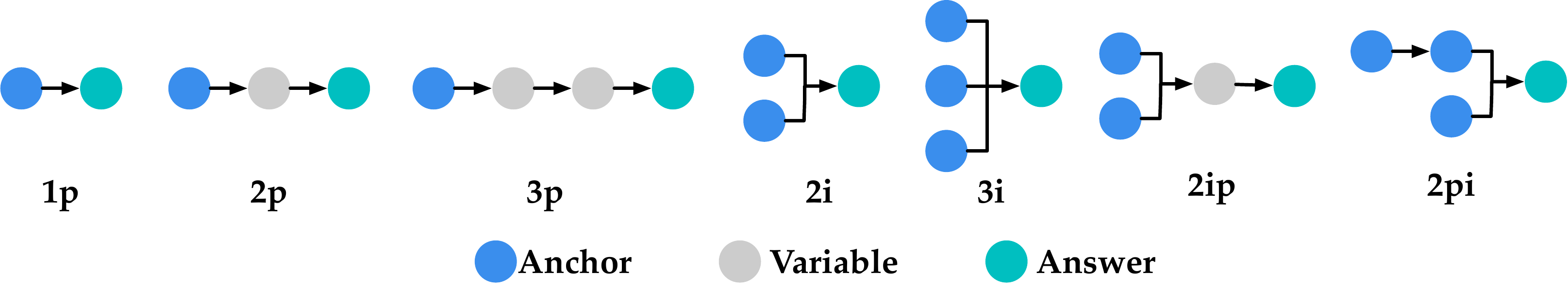, width = 150mm}
    \caption{Templates of query structures, where ``$n$-p'' refers to $n$-hop projection; ``$n$-i'' refers to $n$-way intersection; ``ip'' refers to first intersection then projection; ``pi'' refers to first projection then intersection.}
    \label{fig:qstruc}
\end{figure*}

\section{Evaluation}
\label{sec:expt}

Next, we evaluate \system using real-world multi-view KGs. The evaluation is designed to center around the following questions: 

\vspace{1pt}
\noindent
Q$_1$ -- How effective is \system in answering complex logical queries?

\vspace{1pt}
\noindent
Q$_2$ -- What is the impact of each component on its performance?

\vspace{1pt}
\noindent
Q$_3$ -- How are the training dynamics of \system?

\vspace{1pt}
\noindent
Q$_4$ -- Does \system support different embedding geometries?

\vspace{1pt}
\noindent
Q$_5$ -- Is \system able to process queries from unobserved views?

\subsection{Experimental Setting}

\noindent
\textbf{Dataset  --} We mainly use two real-world KG datasets in our evaluation: Integrated Crisis Early Warning System (ICEWS)\mcite{icews} that represents a collection of political and historical events as \kgs; YAGO\mcite{yago, yago4} that represents a large set of real-world entities from Wikipedia as \kgs.

\begin{table}[!ht]{\small
\renewcommand{\arraystretch}{1.2}
\centering
\setlength{\tabcolsep}{3pt}
{
\begin{tabular}{ccccc}
KG& \#entities & \#relation types & \#edges/facts & \#views \\ 
\hline
\hline
ICWES-Q & \multirow{2}{*}{33,082} & \multirow{2}{*}{532} & \multirow{2}{*}{1,192,484} & 10 \\
ICEWS-M & & & & 28 \\
YAGO & 58,597 & 18 & 721,156 & 17 \\
\hline 
 \end{tabular}
\caption{Statistics of KGs used in the evaluation. \label{tab:kg:stat}}}}
\end{table}

\begin{table*}[!ht]{\small
\renewcommand{\arraystretch}{1.2}
\centering
\setlength{\tabcolsep}{2pt}
\begin{tabular}{ ccccc|cc|cc|cc|cc|cc|ccccc}
 \multirow{2}{*}{Dataset} &  \multirow{2}{*}{Model} & &  \multicolumn{14}{c}{ Query Structure} & & \multicolumn{2}{c}{\multirow{2}{*}{Average}} \\ 
 \cline{4-17} 
  & & &
  \multicolumn{2}{c}{1p} & \multicolumn{2}{c}{2p} & \multicolumn{2}{c}{3p} & 
  \multicolumn{2}{c}{2i} & \multicolumn{2}{c}{3i} &
  \multicolumn{2}{c}{2ip} & \multicolumn{2}{c}{2pi} & & \\
\hline
\hline
  & View-Agnostic & & 30.2 & 70.2 & 19.6 & 43.8 & 10.7 & 43.9 & 20.2 & 46.7 & 25.3 & 49.3 & 41.9 & 70.4 & 18.3 & 47.6 & & 23.7 & 53.1 \\ 
 & TTransE & & 35.3 & 83.4 & 20.1 & 48.3 & 11.0 & 45.8 & \cellcolor{Gray} 25.0 & \cellcolor{Gray}58.5 & \cellcolor{Gray} 30.2 & \cellcolor{Gray} 61.0 & 43.7 & 74.8 & 21.3 & 50.3 & & 26.7 & 60.3 \\ 
  ICEWS-Q & TRESCAL & & 35.3 & 83.0 & 20.6 & 48.5 & 11.0 & 44.4 & 24.9 & 58.3 & 30.1 & 60.2 & 44.3 & 75.2 & 20.7 & 50.9 & & 26.7 & 60.1 \\
 & PosEnc+RGNN+Attn & & 34.4 & 81.2 & 21.6 & 48.8 & 14.3 & 46.9 & 20.4 & 49.9 & 28.8 & 58.8 & 42.4 & 72.0 & 20.8 & 49.2 & & 26.1 & 58.1 \\
  \cline{2-20}
 & \system & & \cellcolor{Gray} 36.8 &\cellcolor{Gray} 84.1 & \cellcolor{Gray} 23.9 & \cellcolor{Gray} 51.1 & \cellcolor{Gray} 15.8 & \cellcolor{Gray} 49.7 & 24.5 & 57.1 & 29.3 & 59.9 & \cellcolor{Gray} 49.2 & \cellcolor{Gray} 80.2 & \cellcolor{Gray} 22.2 & \cellcolor{Gray} 51.8 & \cellcolor{Gray} & \cellcolor{Gray} 28.8 & \cellcolor{Gray} 62.0 \\
\hline 
 & View-Agnostic & & \multicolumn{17}{c}{Same as ICEWS-Q} \\
 & TTransE & & 31.1 & 72.1 & 27.2 & 49.3 & 12.0 & 43.1 & \cellcolor{Gray} 23.1 & \cellcolor{Gray} 54.0 & \cellcolor{Gray} 27.2 & 51.0 & 42.6 & 74.3 & 20.9 & 50.0 & & 26.3 & 56.3 \\ 
 ICEWS-M & TRESCAL & & 31.0 & 72.9 & 27.3 & 49.5 & 12.1 & 44.0 & 23.0 & 53.3 & 26.7 &  \cellcolor{Gray} 51.8 & 42.5 & 72.0 & 20.6 & 49.7 & & 26.2 & 56.2 \\
 & PosEnc+RGNN+Attn & & 30.8 & 68.3 & 25.6 & 45.3 & 14.7 & 44.6 & 19.2 & 49.9 & 22.0 & 46.5 & 44.7 & 74.9 & 21.0 & 48.6 & & 25.4 & 54.0 \\
\cline{2-20}
 & \system & & \cellcolor{Gray} 32.4 & \cellcolor{Gray} 73.6 & \cellcolor{Gray} 29.7 & \cellcolor{Gray} 53.1 & \cellcolor{Gray} 16.0 & \cellcolor{Gray} 45.2 & 22.6 & 53.3 & 26.4 & 49.2 & \cellcolor{Gray} 51.8 & \cellcolor{Gray} 78.8 & \cellcolor{Gray} 22.1 & \cellcolor{Gray} 51.4 & \cellcolor{Gray} & \cellcolor{Gray} 28.7 & \cellcolor{Gray} 57.8 \\
\hline
 \multirow{6}{*}{YAGO} & View-Agnostic & & 36.7 & 77.8 & 18.9 & 89.7 & 10.3 & 76.4 & 32.4 & 49.5 & 70.1 & 91.4 & 88.4 & 95.3 & 20.1 & 30.1 & & 39.6 & 72.9 \\ 
 & TTransE & & 41.5 & 83.7 & 21.6 & 95.2 & 11.2 & 79.1 & 39.2 & 52.9 & 76.5 & 97.2 & 91.7 & 99.3 & 22.6 & 31.7 & & 43.5 & 77.0 \\ 
 & TRESCAL & & 42.0 & 84.5 & 21.4 & 96.0 & 10.1 & 79.6 & 39.7 & 53.3 & 76.5 & 96.8 & 91.2 & 98.9 & 21.8 & 31.4 & & 43.4 & 77.2 \\
 & PosEnc+RGNN+Attn & & 40.2 & 78.7 & 21.8 & 95.5 & 19.1 & 90.8 & 40.5 & 54.4 & 73.5 & 94.2 & 90.4 & 96.9 & 31.3 & 45.8 & & 45.3 & 79.5 \\
\cline{2-20}
 & \system & & \cellcolor{Gray} 46.1 & \cellcolor{Gray} 88.1 & \cellcolor{Gray} 24.3 & \cellcolor{Gray} 97.4 & \cellcolor{Gray} 22.0 & \cellcolor{Gray} 92.9 & \cellcolor{Gray} 44.4 & \cellcolor{Gray} 59.5 & \cellcolor{Gray} 79.8 & \cellcolor{Gray} 97.6 & \cellcolor{Gray} 92.0 & \cellcolor{Gray} 99.6 & \cellcolor{Gray} 32.8 & \cellcolor{Gray} 49.0 & \cellcolor{Gray} & \cellcolor{Gray} 48.8 & \cellcolor{Gray} 83.4 \\
 \end{tabular}}
\caption{Comparison of \system and baselines in answering queries with varying structures (left: \mrr; right: HIT@5)
\label{tab:baseline}}
\end{table*}

For ICEWS, we only keep data reports since 2018 and split them based on month or quarter (3 months). We thus construct two \kgs $\textsf{\small ICWES-Q}$ and $\textsf{\small ICWES-M}$ with varying view granularity. For YAGO, we use the data ranging from 2020 to 2015 to construct the \kg. The statistics of these \kgs are listed in Table\mref{tab:kg:stat}.

Following\mcite{query2box, beta-embedding}, we build queries with different structures as shown in Figure\mref{fig:qstruc}. In training, we sample queries with $\textsf{\small 1p, 2p, 2i, 3i}$ structures from \kgs; at evaluation, we consider queries of all these structures to assess the performance of \system in processing queries with unseen structures. By default, each query with $\textsf{\small 1p, 2p, 3p}$ structures are specified with equal match constraints, while queries with $\textsf{\small 2i, 3i, 2ip, 2pi}$ structures are specified with wildcard match constraints. 
However, when comparing with the baseline methods for temporal KGs, we only sample queries with equal match constraints, given that the baselines cannot handle cross-view queries.

For all the \kgs, we randomly remove half of the relations during generating training samples and evaluate \system's performance under the setting that the relations in the test queries are not present during training.

\vspace{3pt}
\noindent
\textbf{Baselines --} We compare \system with baseline methods on temporal KG learning. While most temporal KRL methods focus on KG completion tasks, we only use their encoders and pair with the same decoder of \system for comparison. We consider the following baselines:
\begin{mitemize}
\item $\textsf{\small TTransE}$\mcite{ttranse}, which extends TransE\mcite{transe} with time-specific parameters. 
Note that TTransE does not assume temporal dependencies for KRL.

\item $\textsf{\small TRESCAL}$\mcite{trescal}, which extends RESCAL\mcite{rescal} with a relation-time-specific weight matrix. 
Like TTransE, TRESCAL also does not assume temporal dependencies for KRL.

\item  $\textsf{\small PosEnc+RGNN+Attn}$, which combines positional encoding, relational GNNs, and self-attention layers as building blocks to encode temporal dynamics of KGs.
A similar model\mcite{iclr2020-temporal-graph} is used for temporal graph learning. \end{mitemize}
The baselines represent two lines of work: one explicitly models the view dynamics with trainable parameters; the other implicitly captures the view dynamics using models while assuming sequential dependencies among the views. Besides, we also consider using \kg without view information (view-agnostic) as a baseline.


\vspace{3pt}
\noindent
\textbf{Metrics --} We evaluate different methods in terms of their ranking of ground-truth answers. We use two popular metrics, \mrr and \hit, for benchmarking KRL methods\mcite{query2box, beta-embedding, temp, temporal-box-embed, kdd-temporal-kgc}: \mrr (mean reciprocal rank) measures the average reciprocal rank of each ground-truth answer; \hit measures the fraction of queries with their ground-truth answers appearing in the top-$K$  candidate answers returned by the methods ($K$ = 5 by default in our evaluation). For both metrics, higher measures indicate better ranking quality.

\vspace{3pt}
\noindent
\textbf{Implementation --} The default implementation of \system consists of a 2-layer, 2-head self-attention encoder and a Query2Box\mcite{query2box} decoder. In \msec{ssec:geo}, we also consider other types of decoders. We train \system with the following setting: optimizer -- Adam\mcite{adam-optimizer}, learning rate -- $10^{-3}$, training steps -- $10^6$, and batch size -- 1,024 samples. 


\subsection{Q1: Baseline Comparison}
\label{ssec:baseline}

In the first set of experiments, we compare \system with baselines in answering logical queries over multi-view KGs. Note that as temporal KRL methods only support queries for snapshots of the KGs (\meg, at a specific time), to make a fair comparison, we thus only consider queries with equal match constraints (\mie, all the relations in the queries appear in the same view). In \msec{ssec:ablation}, we further consider cross-view queries and evaluate \system's performance using an ablation study.

\vspace{3pt}
\textbf{Results --} Table\mref{tab:baseline} summarizes the comparison of \system and other baselines. We have the following observations.

First, it is critical to account for view information in answering logical queries over multi-view KGs. All the other methods, except for  $\textsf{\small PosEnc+RGNN+Attn}$ in the conjunctive structures, significantly outperform \textsf{\small View-Agnostic} KRL.


Second, \system also outperforms the temporal KRL baselines (\mie, $\textsf{\small TTransE}$, $\textsf{\small TRESCAL}$, and $\textsf{\small PosEnc+RGNN+Attn}$) in most cases, especially in the cases of multi-hop query structures (\meg, $\textsf{2p}$, $\textsf{3p}$, $\textsf{2ip}$, $\textsf{2pi}$). This may be explained as follows: the view decoder of \system is optimized to directly enforce view constraints (\meg, equal match); in comparison, $\textsf{\small TTransE}$ and $\textsf{\small TRESCAL}$ attempt to (indirectly) distinguish relations from different views, while $\textsf{\small PosEnc+RGNN+Attn}$ uses models to implicitly encode the view dynamics, resulting in their ineffectiveness of enforcing view constraints.

Finally, the view granularity has a limited impact on \system performance. For instance, comparing the KGs of ICEWS-Q (10 views) and ICEWS-M (28 views), across different query structures, the MRR of \system only slightly drops and even increases in certain cases (\meg, $\textsf{\small 2p}$).

\subsection{Q2: Ablation Study}
\label{ssec:ablation}

In this set of experiments, we conduct an ablation study to understand the impact of each component of \system on its performance. Specifically, we create a set of variants of \system by removing its encoder $\phi$, residual blocks, and view decoder $\ssub{\psi}{\Theta}$, respectively: without the encoder, we directly feed the initial encodings to the decoder; without residual blocks, both relation and view decoders use the embeddings of KG entities as inputs; without view decoder, \system solely relies on the relation decoder. In particular, we measure the performance of such variants in answering cross-view queries, in which the relations of a query are sampled from different views.


\begin{table}[!ht]{\small
\renewcommand{\arraystretch}{1.2}
\centering
\setlength{\tabcolsep}{3pt}
{
\begin{tabular}{ ccc|cc|cc}
\multirow{3}{*}{\system Ablation} & \multicolumn{6}{c}{ Dataset }\\ 
 \cline{2-7} 
  &
  \multicolumn{2}{c}{ICEWS-Q} & \multicolumn{2}{c}{ICEWS-M} & \multicolumn{2}{c}{\multirow{1}{*}{YAGO}} \\
\hline
\hline
w/o encoder & 25.1 & 57.3 & 23.7 & 50.4 & 38.6 & 74.7  \\
w/o residual blocks & 22.7 & 51.3 & 20.2 & 47.8 & 34.4 & 61.9  \\
w/o view decoder & 25.5 & 57.3 & 22.9 & 51.2 & 36.6 & 71.5  \\
\system - full & \cellcolor{Gray} 27.2 & \cellcolor{Gray} 60.1 & \cellcolor{Gray} 24.8 & \cellcolor{Gray} 55.0 & \cellcolor{Gray} 41.3 & \cellcolor{Gray} 77.2  \\
 \end{tabular}
\caption{Ablation study of \system (left: \mrr; right: HIT@5). \label{tab:ablation}}}}
\end{table}

\textbf{Results --} Table\mref{tab:ablation} reports the \mrr and HIT@5 of different variants of \system. As expected, the full-fledged \system attains the best performance across all the datasets. Moreover, observe that removing the residual blocks causes the most significant performance drop. This may be explained by that without the residual blocks, both relation and view decoders use the same latent embeddings as inputs, while the residual blocks help the relation decoder and view decoder focus on the relation encodings and the view encodings, respectively, leading to more effective enforcement of relation and view constraints. Further, the self-attention encoder captures the relational structure of \kgs, while the view decoder enforces the view constraints in the reasoning process, both leading to more effective reasoning under complex relation and view constraints. 

\subsection{Q3: Training Dynamics}
\label{ssec:vermatch}

In this set of experiments, we aim to understand the training dynamics of \system, especially in terms of matching the relation and view constraints of the queries. Recall that for a given query $q$,
\system ranks each candidate answer $a$ by a joint score $\textsf{\small Sim}(a)$ as the product of $ \ssub{\textsf{\small Sim}}{R}(a)$ (\meq{eq:struc:score}) and $\ssub{\textsf{\small Sim}}{\Theta}(a)$ (\meq{eq:ver:score}). To characterize the training dynamics, during the training, we measure (i) the HIT@5 of identifying the ground-truth answers using the joint score and (ii) the HIT@5 of identifying the views of the ground-truth answers using the view score $\ssub{\textsf{\small Sim}}{\Theta}(a)$.



\begin{figure}[!ht]
    \centering
    \epsfig{file = 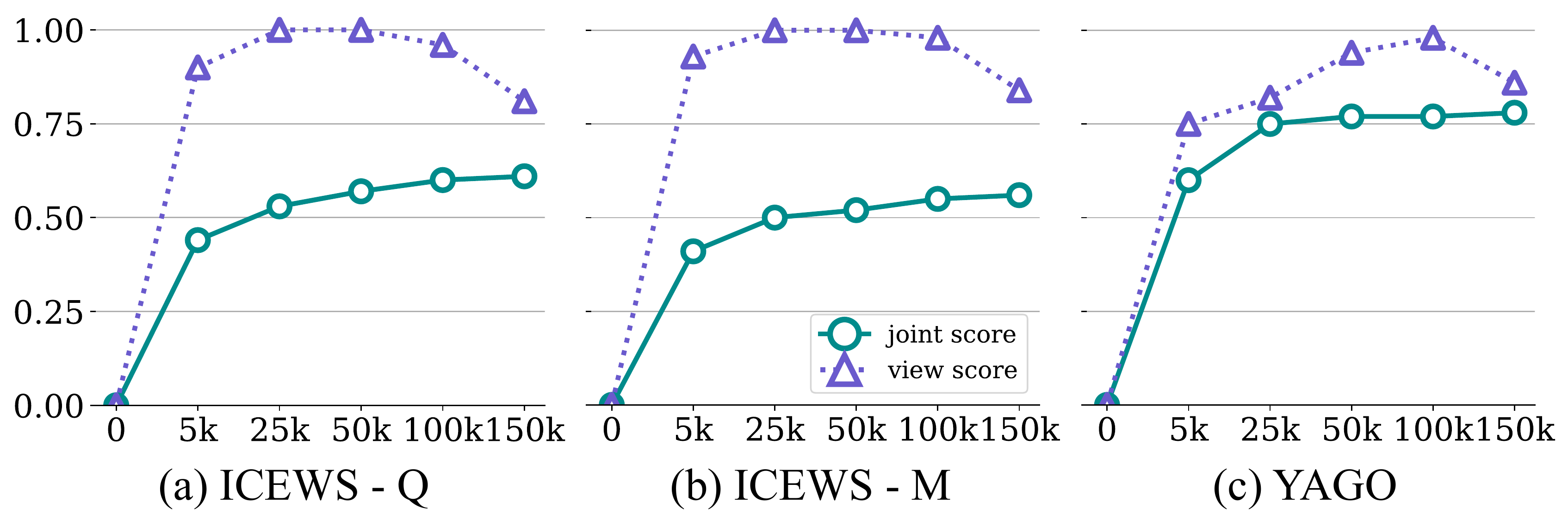, width = 88mm}
    \caption{HIT@5 of \system in identifying the ground-truth entities and the ground-truth views during training (x-axis represents the training steps).}
    \label{fig:vmatch}
\end{figure}

\textbf{Results --} Figure\mref{fig:vmatch} plots the HIT@5 of identifying ground-truth entities and views during the training process. Observe that across all the KGs, \system steadily improves the accuracy of identifying ground-truth answers during 150K steps of training, indicating the effectiveness of training.
In comparison, within the first 5K steps, \system already attains a high accuracy of identifying the correct views, then reaches 1.0 HIT@5 score during the 50K to 100K steps, and slightly drops after that. This may be explained as follows: both relation matching and view matching contribute to identifying the ground-truth answers; yet, it is more challenging to perform relation matching given its more varying structures, which explains the quick increase of view matching accuracy; the overall matching tends to be a compromise between relation and view matching, which explains the slight drop of the view matching accuracy during the late training stage.



\subsection{Q4: Embedding Geometry}
\label{ssec:geo}

We use Query2Box\mcite{query2box} as the default decoder in \system, which projects KG entity embeddings to high-dimensional rectangles (\mie, boxes) in the latent space and models relational and logical operators as transformations over boxes. Besides boxes, there are other possible designs of embedding geometry (\meg, distribution). In this set of experiments, we evaluate the impact of embedding geometry on the performance of \system. Specifically, we consider two other types: vector embeddings and multi-dimensional distributions: GQE\mcite{hamilton2018embedding} represents KG entities as points in the latent space, while BetaE\mcite{beta-embedding} represents KG entities as multi-dimensional Beta distributions in the latent space (using KL-divergence to measure entity distance). We then instantiate the decoder of \system with GQE or BetaE and evaluate its performance.



\begin{figure}[!ht]
    \centering
    \epsfig{file = 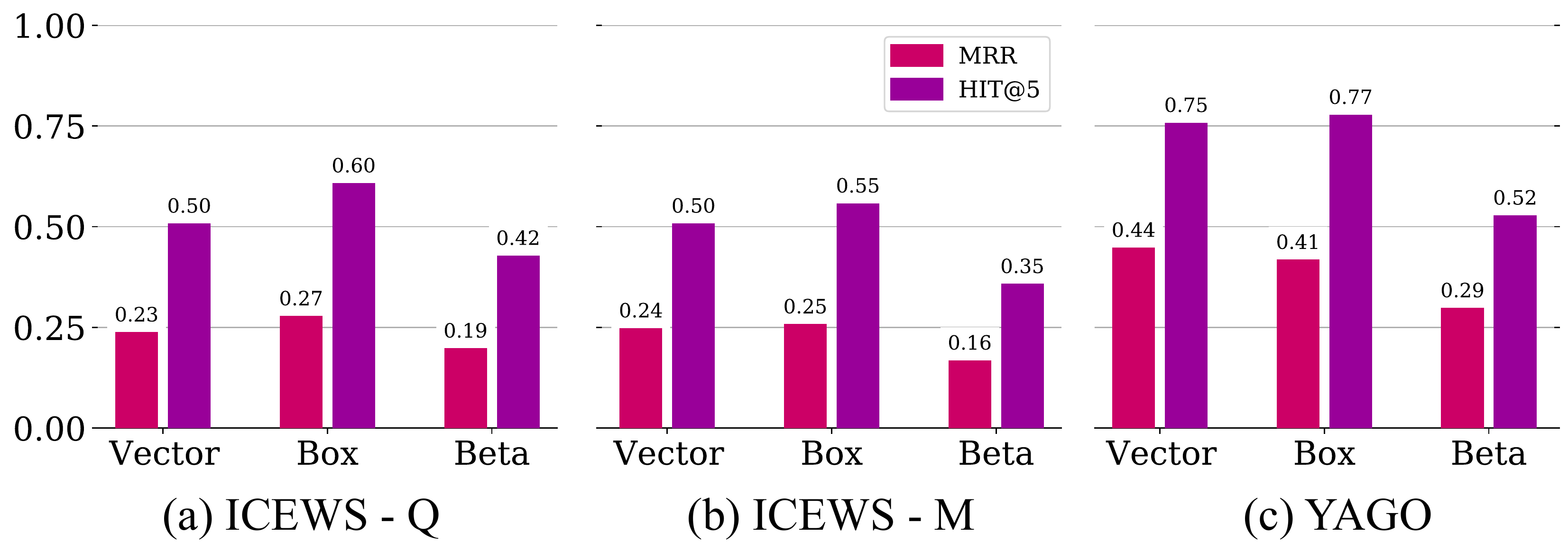, width = 88mm}
    \caption{Performance of \system with different embedding geometries (from left to right: vector\mcite{hamilton2018embedding}, box\mcite{query2box}, and Beta distribution\mcite{beta-embedding})}
    \label{fig:geo}
\end{figure}

\textbf{Results --} We compare the performance (MRR and HIT@5) of \system with different decoders, with results shown in Figure\mref{fig:geo}. Observe that across all the KGs, \system attains similar scores with vector and box embeddings, while it performs slightly worse with distribution embeddings. One possible explanation is that the long tails of Beta distributions may interfere with \system to distinguish relations from different views, while the vector and box embeddings do not suffer such issues.  



\subsection{Q5: Unobserved Views}
\label{ssec:inductive}

All the previous evaluation includes all the views during training; in other words, the test queries are sampled from the views visible to \system. In this set of experiments, we evaluate the performance of \system in answering queries sampled from views that are not included in the training (\mie, unobserved views). To simulate this setting, we define a pivotal view $\vartheta$, we use all the views preceding $\vartheta$ during the training and sample the queries from the views subsequent to $\vartheta$ in evaluation. 



\begin{figure}[!ht]
    \centering
    \epsfig{file = 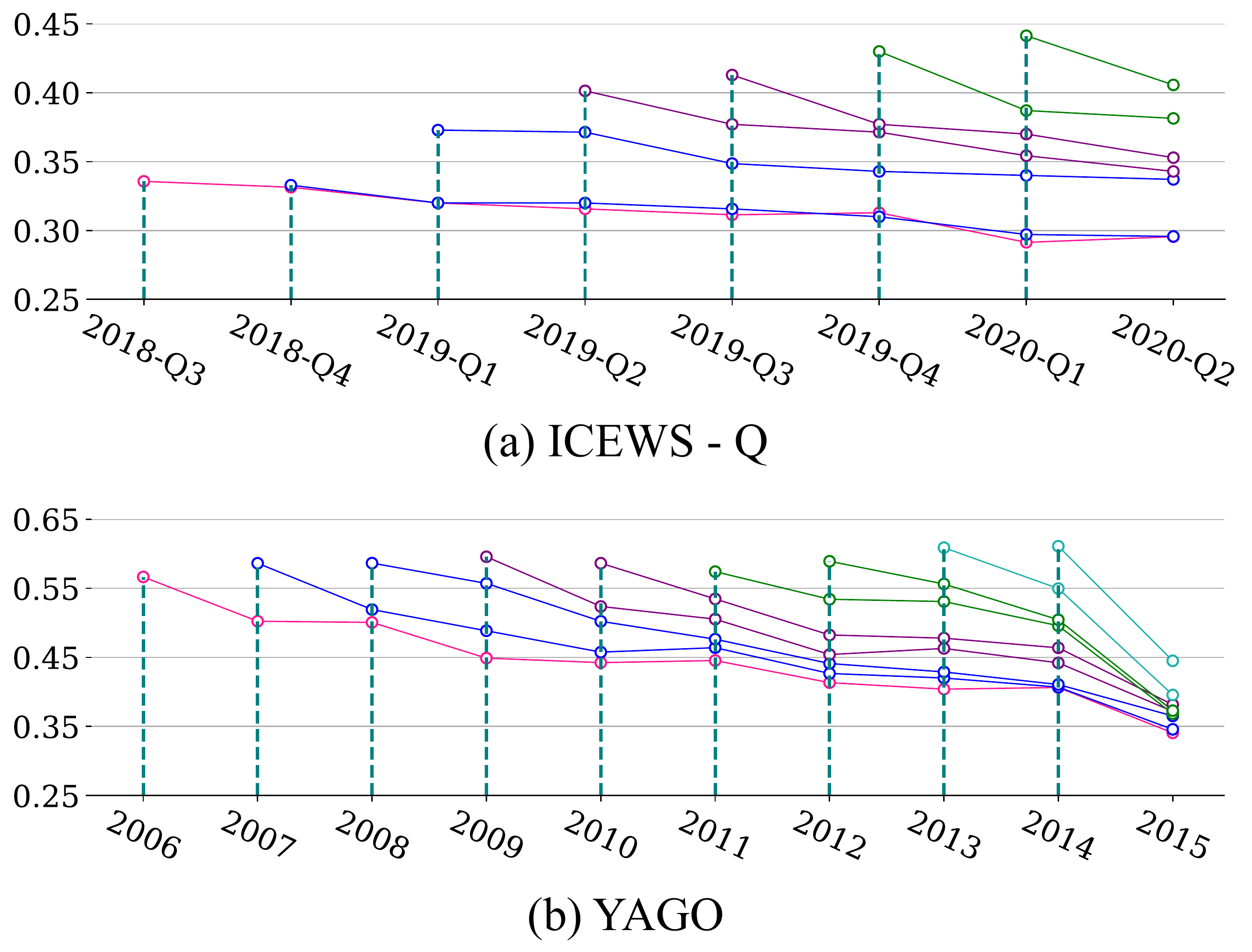, width = 88mm}
    \caption{HIT@5 of \system in reasoning with unobserved views.}
    \label{fig:inductive}
\end{figure}
\textbf{Results --} Figure\mref{fig:inductive} summarizes the performance of \system in answering queries from unobserved views on $\textsf{\small ICEWS-Quarter}$ and $\textsf{\small YAGO}$: the starting point of each line corresponds to the pivotal view $\vartheta$; we then measure the HIT@5 of \system in answering queries sampled from each subsequent view. From Figure\mref{fig:inductive}, we have the following observations. 
First, \system attains reasonable HIT@5 scores (above 0.30 on ICEWS and above 0.35 on YAGO) in answering queries from unobserved views. Second, as expected, the performance of \system tends to drop as the queries are sampled from views distant from the pivotal view. Third, the performance of \system tends to improve as more views are used in the training. These observations indicate that \system is able to capture the dynamics of \kgs that evolve over time, while such dynamics may demonstrate temporal locality. 


\section{Related Work}
\label{sec:literature}

Next, we survey the literature relevant to this work.  

\vspace{3pt}
{\bf Knowledge Representation Learning --} Knowledge graphs (\kgs) represent valuable information sources in various domains\mcite{zhu2020knowledge}. Recent years have witnessed significant progress in using machine learning to reason over \kgs. The existing work can be roughly classified into two categories.
One line of work aims to develop effective \kg embeddings\mcite{transe, transr, rotate, iclr-kge-tricks} such that the semantics of \kg entities/relations are effectively captured by their latent representations for applications such as link prediction\mcite{chains-of-reasoning,path-query}. Another line of work focuses on directly making predictions about complex logical queries\mcite{logic-query-embedding, query2box, beta-embedding} and recommendations\mcite{kgat, kdd-path-recommendation}.
The \krl models considered in this paper belong to this category.


\vspace{3pt}
{\bf Temporal KG Learning --} Besides static KGs, recent studies also investigate KRL for KGs that evolve over time\mcite{chronor, explainable-temporal-kg, tensor-decomp, temporal-krl, temporal-box-embed, temp, kdd-temporal-kgc, temporal-kg-encode, know-evolve}. The fundamental research question is how to effectively encode the temporal dynamics of such KGs.

One approach is to explicitly encode temporal parameters in the latent representations.
For instance, TTransE\mcite{ttranse} extends the score function of TransE\mcite{transe} by adding the time-specific offset or coefficient in the embedding; ChronoR\mcite{chronor} follows RotatE\mcite{rotate} and defines a joint projection function for each relation-time pair; and BoxTE\mcite{temporal-box-embed} applies time-specific projection besides using BoxE\mcite{boxe} for each relation type. 

Another approach is to use neural network models to implicitly capture the temporal dynamics. For instance, one line of work applies recurrent models (\meg, LSTM or GRU) to encode the temporal dynamics\mcite{temporal-kg-encode, temp, temporal-krl}, which consider a temporal KG as a sequence of snapshots and process such snapshots sequentially. Another line of work applies attention mechanisms and propagates attention along with both KG structure and temporal dependency\mcite{kdd-temporal-kgc, explainable-temporal-kg}.

However, it is worth noting that the existing work relies on the temporal dependency among the snapshots of KGs and focuses on the task of KG completion (or link prediction). This work departs in major aspects. (i) By modeling a dynamic KG as a set of overlaying sub-KGs, we lift the assumption of temporal dependency, which allows the definition of multi-view KGs to subsume many different KG types (e.g., static and temporal KGs). (ii) We consider answering complex logical queries that involve varying logical structure across multiple views, which represents an open challenge.


\vspace{3pt}
{\bf Dynamic Graph Learning --} Within a broad context, this work is related to dynamic graph learning, with the aim of modeling the dynamics of temporally evolving graphs. The encoder-decoder framework\mcite{dynamic-gnn-survey} represents one popular design for dynamic graph learning: the encoder aggregates graph structure with temporal dependency to generate latent representations, using network embeddings\mcite{icjai-dne, subset-node-embed}, graph neural networks\mcite{gc-lstm, evolvegcn, icml-dynamic-gnn-rl}, sequential models\mcite{gc-lstm}, and/or self-attention mechanisms\mcite{iclr2020-temporal-graph, twitter-temporal-gnn, dysat}; The decoder maps embeddings towards concrete downstream tasks (\meg, link prediction\mcite{dysat, gc-lstm} and node classification\mcite{evolvegcn, dyane, icjai-dne}). This work is also built upon this encoder-decoder framework.

\section{Conclusion and Future Work}
\label{sec:conclusion}

To our best knowledge, \system represents the first generic reasoning framework for multi-view KGs, which supports answering complex logical queries across multiple views. Extensive empirical evaluation demonstrates that \system significantly outperforms alternative designs in terms of both accuracy and scalability. This work also opens up several directions for further investigation: (i) support of queries beyond first-order conjunctive queries (\meg, adjunctive queries); (ii) incorporation of semantic correlation across different views; (iii) exploration of designs other than the encoder-decoder framework; and (iv) evaluation on non-temporal dynamic KG datasets.


\bibliographystyle{plain}
\bibliography{ref.bib}

\begin{thebibliography}{10}

\bibitem{yago}
{YAGO: A High-Quality Knowledge Base}.
\newblock \url{https://yago-knowledge.org/}.

\bibitem{boxe}
Ralph Abboud, Ismail~Ilkan Ceylan, Thomas Lukasiewicz, and Tommaso Salvatori.
\newblock {Boxe: A Box Embedding Model for Knowledge Base Completion}.
\newblock {\em ArXiv e-prints}, 2020.

\bibitem{transe}
Antoine Bordes, Nicolas Usunier, Alberto Garcia-Dur\'{a}n, Jason Weston, and
  Oksana Yakhnenko.
\newblock {Translating Embeddings for Modeling Multi-Relational Data}.
\newblock In {\em Proceedings of Advances in Neural Information Processing
  Systems (NeurIPS)}, 2013.

\bibitem{icews}
E~Boschee, J~Lautenschlager, S~O’Brien, S~Shellman, J~Starz, and M~Ward.
\newblock {Integrated Crisis Early Warning System (ICEWS) Coded Event Data}.
\newblock {\em URL: https://dataverse.harvard.edu/dataverse/icews}, 2015.

\bibitem{gc-lstm}
Jinyin Chen, Xuanheng Xu, Yangyang Wu, and Haibin Zheng.
\newblock {Gc-lstm: Graph Convolution Embedded LSTM for Dynamic Link
  Prediction}.
\newblock {\em ArXiv e-prints}, 2018.

\bibitem{multilingual}
Muhao Chen, Yingtao Tian, Mohan Yang, and Carlo Zaniolo.
\newblock {Multilingual Knowledge Graph Embeddings for Cross-lingual Knowledge
  Alignment}.
\newblock {\em ArXiv e-prints}, 2016.

\bibitem{missing-relation}
Nilesh Dalvi and Dan Suciu.
\newblock {Efficient Query Evaluation on Probabilistic Databases}.
\newblock {\em The VLDB Journal}, 16:523--544, 2007.

\bibitem{path-query}
Rajarshi {Das}, Shehzaad {Dhuliawala}, Manzil {Zaheer}, Luke {Vilnis}, Ishan
  {Durugkar}, Akshay {Krishnamurthy}, Alex {Smola}, and Andrew {McCallum}.
\newblock {Go for a Walk and Arrive at the Answer: Reasoning Over Paths in
  Knowledge Bases using Reinforcement Learning}.
\newblock In {\em Proceedings of International Conference on Learning
  Representations (ICLR)}, 2018.

\bibitem{chains-of-reasoning}
Rajarshi {Das}, Arvind {Neelakantan}, David {Belanger}, and Andrew {McCallum}.
\newblock {Chains of Reasoning over Entities, Relations, and Text using
  Recurrent Neural Networks}.
\newblock In {\em Proceedings of European Chapter of the Association for
  Computational Linguistics (EACL)}, 2017.

\bibitem{icjai-dne}
Lun Du, Yun Wang, Guojie Song, Zhicong Lu, and Junshan Wang.
\newblock {Dynamic Network Embedding: An Extended Approach for Skip-gram based
  Network Embedding.}
\newblock In {\em Proceedings of International Joint Conference on Artificial
  Intelligence (IJCAI)}, 2018.

\bibitem{temporal-kg-encode}
Alberto Garc{\'\i}a-Dur{\'a}n, Sebastijan Duman{\v{c}}i{\'c}, and Mathias
  Niepert.
\newblock {Learning Sequence Encoders for Temporal Knowledge Graph Completion}.
\newblock {\em ArXiv e-prints}, 2018.

\bibitem{subset-node-embed}
Xingzhi Guo, Baojian Zhou, and Steven Skiena.
\newblock {Subset Node Representation Learning over Large Dynamic Graphs}.
\newblock {\em ArXiv e-prints}, 2021.

\bibitem{logic-query-embedding}
William~L. {Hamilton}, Payal {Bajaj}, Marinka {Zitnik}, Dan {Jurafsky}, and
  Jure {Leskovec}.
\newblock {Embedding Logical Queries on Knowledge Graphs}.
\newblock In {\em Proceedings of Advances in Neural Information Processing
  Systems (NeurIPS)}, 2018.

\bibitem{hamilton2018embedding}
William~L Hamilton, Payal Bajaj, Marinka Zitnik, Dan Jurafsky, and Jure
  Leskovec.
\newblock {Embedding Logical Queries on Knowledge Graphs}.
\newblock {\em ArXiv e-prints}, 2018.

\bibitem{explainable-temporal-kg}
Zhen Han, Peng Chen, Yunpu Ma, and Volker Tresp.
\newblock {Explainable Subgraph Reasoning for Forecasting on Temporal Knowledge
  Graphs}.
\newblock In {\em Proceedings of International Conference on Learning
  Representations (ICLR)}, 2020.

\bibitem{kdd-temporal-kgc}
Jaehun Jung, Jinhong Jung, and U~Kang.
\newblock {Learning to Walk Across Time for Interpretable Temporal Knowledge
  Graph Completion}.
\newblock In {\em Proceedings of ACM International Conference on Knowledge
  Discovery and Data Mining (KDD)}, 2021.

\bibitem{dynamic-gnn-survey}
Seyed~Mehran Kazemi, Rishab Goel, Kshitij Jain, Ivan Kobyzev, Akshay Sethi,
  Peter Forsyth, and Pascal Poupart.
\newblock {Representation Learning for Dynamic Graphs: A Survey.}
\newblock {\em J. Mach. Learn. Res.}, 21:1--73, 2020.

\bibitem{adam-optimizer}
Diederik~P Kingma and Jimmy Ba.
\newblock {Adam: A method for Stochastic Optimization}.
\newblock {\em ArXiv e-prints}, 2014.

\bibitem{tensor-decomp}
Timoth{\'e}e Lacroix, Guillaume Obozinski, and Nicolas Usunier.
\newblock {Tensor Decompositions for Temporal Knowledge Base Completion}.
\newblock {\em Proceedings of International Conference on Learning
  Representations (ICLR)}, 2020.

\bibitem{ttranse}
Julien Leblay and Melisachew~Wudage Chekol.
\newblock Deriving validity time in knowledge graph.
\newblock In {\em Companion Proceedings of the The Web Conference 2018}, pages
  1771--1776, 2018.

\bibitem{kdd-path-recommendation}
Wenqiang Lei, Gangyi Zhang, Xiangnan He, Yisong Miao, Xiang Wang, Liang Chen,
  and Tat-Seng Chua.
\newblock {Interactive Path Reasoning on Graph for Conversational
  Recommendation}.
\newblock In {\em Proceedings of ACM International Conference on Knowledge
  Discovery and Data Mining (KDD)}, 2020.

\bibitem{temporal-krl}
Zixuan Li, Xiaolong Jin, Wei Li, Saiping Guan, Jiafeng Guo, Huawei Shen,
  Yuanzhuo Wang, and Xueqi Cheng.
\newblock {Temporal Knowledge Graph Reasoning Based on Evolutional
  Representation Learning}.
\newblock {\em ArXiv e-prints}, 2021.

\bibitem{transr}
Yankai Lin, Zhiyuan Liu, Maosong Sun, Yang Liu, and Xuan Zhu.
\newblock {Learning Entity and Relation Embeddings for Knowledge Graph
  Completion}.
\newblock In {\em Proceedings of AAAI Conference on Artificial Intelligence
  (AAAI)}, 2015.

\bibitem{cross-lingual-kg}
Xin Mao, Wenting Wang, Huimin Xu, Man Lan, and Yuanbin Wu.
\newblock {MRAEA: an Efficient and Robust Entity Alignment Approach for
  Cross-lingual Knowledge Graph}.
\newblock In {\em Proceedings of the International Conference on Web Search and
  Data Mining}, 2020.

\bibitem{icml-dynamic-gnn-rl}
Eli Meirom, Haggai Maron, Shie Mannor, and Gal Chechik.
\newblock {Controlling Graph Dynamics with Reinforcement Learning and Graph
  Neural Networks}.
\newblock In {\em Proceedings of IEEE Conference on Machine Learning (ICML)},
  2021.

\bibitem{temporal-box-embed}
Johannes Messner, Ralph Abboud, and Ismail~Ilkan Ceylan.
\newblock {Temporal Knowledge Graph Completion using Box Embeddings}.
\newblock {\em ArXiv e-prints}, 2021.

\bibitem{rescal}
Maximilian Nickel, Volker Tresp, and Hans-Peter Kriegel.
\newblock {A Three-way Model for Collective Learning on Multi-relational Data}.
\newblock In {\em Proceedings of IEEE Conference on Machine Learning (ICML)},
  2011.

\bibitem{evolvegcn}
Aldo Pareja, Giacomo Domeniconi, Jie Chen, Tengfei Ma, Toyotaro Suzumura,
  Hiroki Kanezashi, Tim Kaler, Tao Schardl, and Charles Leiserson.
\newblock {Evolvegcn: Evolving Graph Convolutional Networks for Dynamic
  Graphs}.
\newblock In {\em Proceedings of AAAI Conference on Artificial Intelligence
  (AAAI)}, 2020.

\bibitem{yago4}
Thomas Pellissier~Tanon, Gerhard Weikum, and Fabian Suchanek.
\newblock {Yago 4: A Reason-able Knowledge Base}.
\newblock In {\em European Semantic Web Conference}, pages 583--596. Springer,
  2020.

\bibitem{ren2021lego}
Hongyu Ren, Hanjun Dai, Bo~Dai, Xinyun Chen, Michihiro Yasunaga, Haitian Sun,
  Dale Schuurmans, Jure Leskovec, and Denny Zhou.
\newblock {LEGO: Latent Execution-Guided Reasoning for Multi-Hop Question
  Answering on Knowledge Graphs}.
\newblock In {\em Proceedings of IEEE Conference on Machine Learning (ICML)},
  2021.

\bibitem{query2box}
Hongyu {Ren}, Weihua {Hu}, and Jure {Leskovec}.
\newblock {Query2box: Reasoning over Knowledge Graphs in Vector Space using Box
  Embeddings}.
\newblock In {\em Proceedings of International Conference on Learning
  Representations (ICLR)}, 2020.

\bibitem{beta-embedding}
Hongyu {Ren} and Jure {Leskovec}.
\newblock {Beta Embeddings for Multi-Hop Logical Reasoning in Knowledge
  Graphs}.
\newblock In {\em Proceedings of Advances in Neural Information Processing
  Systems (NeurIPS)}, 2020.

\bibitem{twitter-temporal-gnn}
Emanuele Rossi, Ben Chamberlain, Fabrizio Frasca, Davide Eynard, Federico
  Monti, and Michael Bronstein.
\newblock {Temporal Graph Networks for Deep Learning on Dynamic Graphs}.
\newblock {\em ArXiv e-prints}, 2020.

\bibitem{iclr-kge-tricks}
Daniel Ruffinelli, Samuel Broscheit, and Rainer Gemulla.
\newblock You can teach an old dog new tricks! on training knowledge graph
  embeddings.
\newblock In {\em Proceedings of International Conference on Learning
  Representations (ICLR)}, 2019.

\bibitem{chronor}
Ali Sadeghian, Mohammadreza Armandpour, Anthony Colas, and Daisy~Zhe Wang.
\newblock {ChronoR: Rotation Based Temporal Knowledge Graph Embedding}.
\newblock {\em ArXiv e-prints}, 2021.

\bibitem{dysat}
Aravind Sankar, Yanhong Wu, Liang Gou, Wei Zhang, and Hao Yang.
\newblock {Dynamic Graph Representation Learning via Self-attention Networks}.
\newblock {\em Proceedings of International Conference on Learning
  Representations (ICLR)}, 2019.

\bibitem{dyane}
Koya Sato, Mizuki Oka, Alain Barrat, and Ciro Cattuto.
\newblock Dyane: dynamics-aware node embedding for temporal networks.
\newblock {\em arXiv preprint arXiv:1909.05976}, 2019.

\bibitem{rotate}
Zhiqing Sun, Zhi-Hong Deng, Jian-Yun Nie, and Jian Tang.
\newblock {Rotate: Knowledge Graph Embedding by Relational Rotation in Complex
  Space}.
\newblock {\em Proceedings of International Conference on Learning
  Representations (ICLR)}, 2019.

\bibitem{know-evolve}
Rakshit Trivedi, Hanjun Dai, Yichen Wang, and Le~Song.
\newblock {Know-evolve: Deep Temporal Reasoning for Dynamic Knowledge Graphs}.
\newblock In {\em Proceedings of IEEE Conference on Machine Learning (ICML)},
  2017.

\bibitem{attn-is-all-you-need}
Ashish Vaswani, Noam Shazeer, Niki Parmar, Jakob Uszkoreit, Llion Jones,
  Aidan~N Gomez, {\L}ukasz Kaiser, and Illia Polosukhin.
\newblock {Attention is All You Need}.
\newblock {\em Proceedings of Advances in Neural Information Processing Systems
  (NeurIPS)}, 2017.

\bibitem{gat}
Petar {Veli{\v{c}}kovi{\'c}}, Guillem {Cucurull}, Arantxa {Casanova}, Adriana
  {Romero}, Pietro {Li{\`o}}, and Yoshua {Bengio}.
\newblock {Graph Attention Networks}.
\newblock In {\em Proceedings of International Conference on Learning
  Representations (ICLR)}, 2018.

\bibitem{trescal}
Quan Wang, Bin Wang, and Li~Guo.
\newblock {Knowledge Base Completion Using Embeddings and Rules}.
\newblock In {\em Proceedings of AAAI Conference on Artificial Intelligence
  (AAAI)}, 2015.

\bibitem{kgat}
Xiang Wang, Xiangnan He, Yixin Cao, Meng Liu, and Tat-Seng Chua.
\newblock {Kgat: Knowledge Graph Attention Network for Recommendation}.
\newblock In {\em Proceedings of ACM International Conference on Knowledge
  Discovery and Data Mining (KDD)}, 2019.

\bibitem{temp}
Jiapeng Wu, Meng Cao, Jackie Chi~Kit Cheung, and William~L Hamilton.
\newblock {TeMP: Temporal Message Passing for Temporal Knowledge Graph
  Completion}.
\newblock {\em ArXiv e-prints}, 2020.

\bibitem{multi-view-survey}
Chang Xu, Dacheng Tao, and Chao Xu.
\newblock {A Survey on Multi-view Learning}.
\newblock {\em ArXiv e-prints}, 2013.

\bibitem{iclr2020-temporal-graph}
Da~Xu, Chuanwei Ruan, Evren Korpeoglu, Sushant Kumar, and Kannan Achan.
\newblock {Inductive Representation Learning on Temporal Graphs}.
\newblock {\em Proceedings of International Conference on Learning
  Representations (ICLR)}, 2020.

\bibitem{deepset}
Manzil Zaheer, Satwik Kottur, Siamak Ravanbakhsh, Barnabas Poczos, Russ~R
  Salakhutdinov, and Alexander~J Smola.
\newblock {Deep Sets}.
\newblock {\em Proceedings of Advances in Neural Information Processing Systems
  (NeurIPS)}, 2017.

\bibitem{zhu2020knowledge}
Yongjun Zhu, Chao Che, Bo~Jin, Ningrui Zhang, Chang Su, and Fei Wang.
\newblock {Knowledge-driven Drug Repurposing Using a Comprehensive Drug
  Knowledge Graph}.
\newblock {\em Health Informatics Journal}, 26, 2020.

\end{thebibliography}


\end{document}